\documentclass[conference]{IEEEtran}
\usepackage[utf8]{inputenc}
\usepackage[T1]{fontenc}
\usepackage{amsmath,amssymb}
\usepackage{graphicx}
\usepackage{booktabs}
\usepackage{cite}
\usepackage[hidelinks]{hyperref}

\newcommand{\Fobj}{F}
\newcommand{\Ftwo}{F_2}
\newcommand{\Gfit}{G}
\newcommand{\Fhat}{\hat{F}}
\DeclareMathOperator*{\argmax}{arg\,max}

\begin{document}

\title{Pairwise Approximation Can Select\\the Wrong Multi-Robot Plan}

\author{\IEEEauthorblockN{William Teo}
\IEEEauthorblockA{MARMoT Lab, Department of Mechanical Engineering,
National University of Singapore\\
NEAR Lab, AI.Robotics Strategic Technology Centre, Singapore Technologies Engineering\\
williamteo@u.nus.edu}}

\IEEEoverridecommandlockouts
\IEEEaftertitletext{%
\begin{center}
\includegraphics[width=\textwidth]{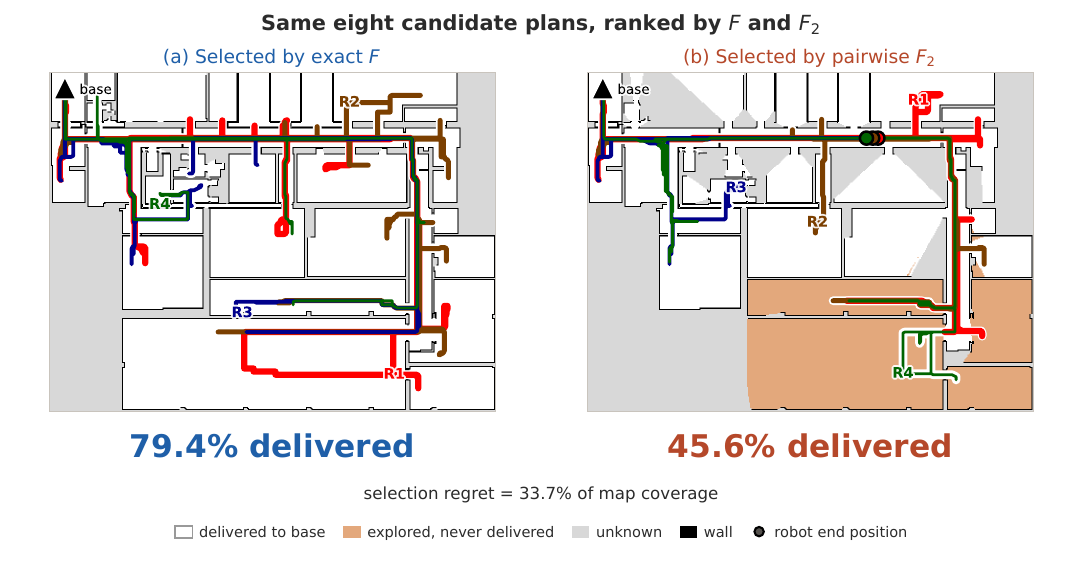}
\refstepcounter{figure}\label{fig:env1}
\parbox{\textwidth}{\footnotesize
\textbf{Fig.~\thefigure. Plans selected on env1 at the 15\,m generation range.}
Both scores rank the same eight logged candidates. Ranking by $\Fobj$ selects the plan in (a), which delivers 79.4\% of the map. Ranking by $\Ftwo$ selects the plan in (b), which delivers 45.6\%, giving 0.337 regret. The warm shaded region in (b) was explored but had not reached the base by episode end. Colored lines show robot trajectories; circles mark final positions away from the base in (b), where R1, R2, and R4 end together in the top corridor and R3 ends near the base. In (a), every robot ends at the base.}
\end{center}}
\maketitle

\begin{abstract}
Multi-robot coordination methods often score a joint plan from singleton and pairwise terms, leaving out the terms that involve three or more robots. We measure the plan-selection regret of two pairwise approximations to delivered coverage using frozen multi-robot trajectories. For each four-robot plan on an indoor exploration benchmark, replaying all 16 robot subsets gives the exact delivered-coverage set function $\Fobj$. From the same subset values we compute two pairwise scores: the exact order-2 M\"obius truncation $\Ftwo$, which depends only on the singleton and pair values, and an equal-weight least-squares two-additive fit $\Gfit$. Ranking by $\Ftwo$ instead of $\Fobj$ changes the selected plan on six of seven maps at the 15\,m candidate-generation range in each of two candidate families, with regret up to 0.337 of map coverage. Switching to $\Gfit$ reduces the regret but still changes the selection on three of seven maps in each family. The additive score $F_1$, which keeps only the singleton terms, selects the exact winner on six of seven maps in one family and four of seven in the other, against one of seven for $\Ftwo$. We also find that lower average reconstruction error does not guarantee lower selection regret.
\end{abstract}

\section{Introduction}

We test whether a score that keeps only singleton and pairwise interaction terms selects the same multi-robot plan as the exact delivered-coverage objective. Figure~\ref{fig:env1} shows a case where the two scores select different plans from the same pool. The plan selected by the pairwise score leaves much of its explored area undelivered to the base station.

We compare exact order-2 truncation~$\Ftwo$ with a least-squares two-additive fit~$\Gfit$ and the additive score~$F_1$. The comparison measures the coverage lost when each score selects a plan from a fixed candidate pool. We also examine whether lower average reconstruction error corresponds to lower selection regret.

For a set function over $n$ robots, restricting its interaction terms to singletons and pairs reduces their number from $2^n-1$ to $n+\binom{n}{2}$. Pairwise interaction terms are common in multi-robot coordination. For example, edge-based coordination graphs~\cite{kok2006collaborative} and pairwise payoff functions in deep multi-agent learning~\cite{bohmer2020deep} keep only singleton and pairwise terms. In coordinated frontier exploration, Burgard et al.~\cite{burgard2005coordinated} reduce the utility of a frontier target by a pairwise visibility penalty for each target already assigned to another robot, so each assignment is scored by its own utility minus pairwise discounts, less a path cost. Other frontier and sequential greedy methods simplify the search rather than the objective, for example by steering robots toward map frontiers or adding robots by marginal gain~\cite{yamauchi1997frontier,yamauchi1998multirobot,krause2008near,singh2009efficient,corah2019distributed}. Factored formulations write the joint value as a sum of terms over small cliques~\cite{guestrin2001multiagent}. A set function whose M\"obius terms vanish above order $k$ is called $k$-additive~\cite{grabisch1997kadditive}; the fitted model below is the two-additive case.

Those methods factor value over joint actions or target assignments. Here we apply the same pairwise restriction to a set function over robot subsets, and measure whether removing higher-order terms changes the selected plan. Worst-case analysis shows that maximizing a submodular function from evaluations on bounded-size subsets alone can be far from optimal~\cite{downie2023submodular}. Under unconstrained communication, delivered coverage is a coverage function, and its order-3 and order-4 M\"obius terms are the signed areas seen jointly by three or by four robots. Under finite communication, delivery can also depend on relay and connectivity patterns involving several robots~\cite{rooker2007multirobot,banfi2018strategies,hull2025bandwidth,kim2026proid}.

We use the indoor exploration benchmark from the IROS 2026 Intelligent Information Gathering workshop~\cite{byufrost2026indoor}, with four-robot teams on its seven released maps. Replaying logged trajectories over all $2^4$ robot subsets evaluates the delivered-coverage set function exactly. The $2^n$ subset replays per candidate are the raw measurements from which the M\"obius decomposition is computed; they are used here for offline analysis, not as a proposed runtime method.

\section{Plan Selection with an Approximate Score}
\label{sec:condition}

We select one plan from a pool $B$. Each candidate is a frozen set of trajectories for the full team $N$. The exact objective assigns each candidate $b$ a full-team score $\Fobj_b(N)$, and a selector picks $\argmax_{b \in B} \Fobj_b(N)$. A pairwise representation replaces $\Fobj$ with an approximate score $\Fhat$ and picks $\argmax_{b \in B} \Fhat_b(N)$ instead. The regret of the approximate selection is $\Fobj_{b_F}(N) - \Fobj_{b'}(N) \ge 0$, where $b_F$ is the plan $\Fobj$ ranks first and $b'$ is the plan $\Fhat$ selects, measured in map coverage. For each score, the top set contains all candidates within $10^{-12}$ of its maximum. Among candidates in the approximate top set, we report the smallest exact regret. We count a selection change when the top sets under the two scores are disjoint or the regret exceeds $10^{-12}$.

Large errors can be harmless if they shift competing plans by about the same amount. A ranking flips only when the difference between those shifts exceeds the exact-score gap. Let $e_b = \Fobj_b(N) - \Fhat_b(N)$ denote a candidate's approximation error, let $w$ be the plan ranked first by $\Fobj$, and let $c$ be a competitor. Apart from ties within the stated tolerance, $c$ outranks $w$ under $\Fhat$ exactly when
\begin{equation}
e_w - e_c \;>\; \Fobj_w(N) - \Fobj_c(N).
\label{eq:condition}
\end{equation}
If all candidates have the same approximation error, every ranking is preserved. Equation~\eqref{eq:condition} uses the exact-score gap between $w$ and $c$. We call the gap between $w$ and the exact runner-up the top-two margin.

Equation~\eqref{eq:condition} compares the errors of two candidates. Averaging errors across the pool hides that difference, so average reconstruction error need not track selection changes. Section~\ref{sec:e2null} compares the two on the benchmark.

Equation~\eqref{eq:condition} is an algebraic identity. We use it to describe observed ranking changes. Because it requires the exact scores, it is not a predictive or runtime test. A selection guarantee based on this identity would need bounds on individual candidates' errors relative to the score gaps between plans. We do not derive such bounds in this paper.

\section{Delivered Coverage and Its Pairwise Representations}
\label{sec:formulation}

\subsection{Exact subset values by frozen replay}
\label{sec:replay}

A team $N = \{1,\dots,n\}$ of robots explores an occupancy-grid indoor environment. Robots sense locally, exchange maps with peers within communication range, subject to the benchmark's wall-attenuated path-loss model, and deliver fused maps to a fixed base station. The benchmark scores an episode by \emph{delivered coverage}: the fraction of the map known to the base station when the episode ends \cite{byufrost2026indoor}.

For a subset $S \subseteq N$ we define $\Fobj(S)$ as the delivered coverage of the episode replayed with only the robots in $S$ present, each following its logged trajectory exactly; absent robots never sense, communicate, or relay. Each replay starts from a fresh world, so evaluation order cannot leak state, and replay never invokes a planning policy. Although the remaining robots do not replan, the subset values still measure how each robot's presence changes which observations reach the base through the benchmark's communication model. $\Fobj(\emptyset) = 0$. Evaluating~$\Fobj$ at all $2^n$ subsets gives every interaction term in Section~\ref{sec:mobius} exactly, including the higher-order terms that the pairwise scores omit. We verify that, under the conditions used to generate each plan, the full-team replay $\Fobj(N)$ reproduces the original score bit-exactly.

\subsection{Exact truncation and fitted approximation}
\label{sec:mobius}

\begin{figure}[t]
\centering
\includegraphics[width=\columnwidth]{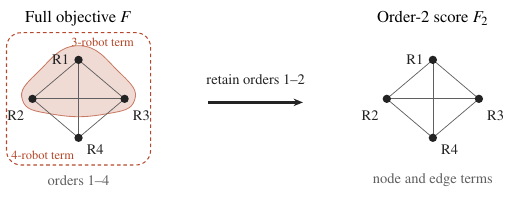}
\caption{Graph view of the interaction-order decomposition. Singleton terms attach to individual robots and pairwise terms to edges. The order-2 score $\Ftwo$ retains these terms and omits the terms involving three or four robots.}
\label{fig:concept}
\end{figure}

In the graph view of Figure~\ref{fig:concept}, singleton terms attach to nodes and pairwise terms to edges. Higher-order terms involve larger robot groups. The M\"obius decomposition makes this split exact \cite{rota1964foundations,harsanyi1963simplified,grabisch1999axiomatic}: every finite set function decomposes into interaction terms
\begin{equation}
m(T) = \sum_{U \subseteq T} (-1)^{|T \setminus U|}\, \Fobj(U),
\qquad
\Fobj(S) = \sum_{T \subseteq S} m(T),
\end{equation}
where $m(T)$ is the part of the value that belongs to the group $T$ jointly and to no smaller group inside it. Keeping only terms up to order two gives the order-2 score
\begin{equation}
\Ftwo(S) = \sum_{T \subseteq S,\; |T| \le 2} m(T),
\end{equation}
the exact sum of the node and edge terms. $\Ftwo$ depends only on the singleton and pair values. For $n = 4$, $\Ftwo(N) = \sum_{i<j} \Fobj(\{i,j\}) - 2\sum_{i} \Fobj(\{i\})$, so it is the team score that a method evaluating only individual robots and pairs assigns by inclusion--exclusion. Keeping only the singleton terms gives the additive score $F_1(S) = \sum_{i \in S} \Fobj(\{i\})$. For $n = 4$, $\Ftwo$ omits five interaction terms: the four triples and the quadruple. The order-4 truncation equals $\Fobj$ identically, which we verify to $10^{-12}$ in every episode.

$\Ftwo$ keeps the original singleton and pairwise terms, whereas a fitted model refits them to the observed subset values. We therefore also evaluate the two-additive fit
\begin{equation}
\Gfit(S) = \sum_{i \in S} a_i \;+\; \sum_{\{i,j\} \subseteq S} b_{ij},
\label{eq:gfit}
\end{equation}
with coefficients chosen by ordinary unregularized least squares. We fit $\Gfit$ separately for each candidate using all 15 non-empty subset values with equal weight, including the full-team value $\Fobj(N)$. The fit has no hyperparameters and no intercept, so $\Gfit(\emptyset) = 0$. For four robots this fits 10 coefficients to 15 equations. Unlike $\Ftwo$, $\Gfit$ uses the triple and quadruple subset values, which a method that evaluates only singletons and pairs does not have. We use $\Ftwo$ and $\Gfit$ as the two pairwise scores in Section~\ref{sec:condition}, and $F_1$ as the additive reference.

The residual $\Fobj(S) - \Ftwo(S)$ is zero for $|S| \le 2$ by construction, so any nonzero residual is order-three-or-four structure. Unlike $\Fobj$, which is a coverage fraction in $[0,1]$, $\Ftwo$ need not stay in this range: removing higher-order terms can push $\Ftwo(S)$ below 0 or above 1.

We summarize reconstruction quality by the normalized error
\begin{equation}
E_2 =
\frac{\sum_{\emptyset \neq S \subseteq N} \bigl|\Fobj(S) - \Ftwo(S)\bigr|}
     {\sum_{\emptyset \neq S \subseteq N} \Fobj(S)}.
\label{eq:e2}
\end{equation}
$E_2$ averages reconstruction error over all subset values. Selection uses the ordering of the full-team candidate scores (Section~\ref{sec:condition}).

\section{Experimental Setup}
\label{sec:protocol}

\textbf{Benchmark.} All experiments use the workshop's indoor exploration competition simulator \cite{byufrost2026indoor}, pinned at commit \texttt{2a2c751}, with its seven released maps (env1--env7). The benchmark's multi-agent track supports teams of two to five robots. We fix the team size at four so that all $2^4 = 16$ robot subsets can be replayed exhaustively. Candidate plans use the shipped local-development start pose $[15,15]$, start delays of 0, 5, 10, and 15 steps, and a 1000-step horizon on every map. Candidate generation uses the benchmark's 15\,m communication-range parameter.

\textbf{Candidate plans.} The primary pool contains eight candidate joint plans per map, generated by the benchmark's shipped nearest-frontier policy under a parameter grid: relay period $\{100, 300\}$ $\times$ relay handoff $\{$on, off$\}$ $\times$ frontier-crowding penalty $\{$default, off$\}$. A second family of eight candidate plans per map replaces the heuristic with uniform-random frontier selection, combining relay period and handoff toggles across generator seeds 0 and 1. All eight resulting trajectories were distinct on every evaluated map. All candidates were generated at the 15\,m range, and all replays are deterministic.

\textbf{Communication conditions.} Each primary-family candidate is replayed at ranges 3, 6, 9, 15, and 30\,m and at an idealized condition with unconstrained communication, under which delivered coverage equals the geometric union of what the team observed. The plans were generated at 15\,m and replayed unchanged at every other range and under idealized communication. The logged trajectories are identical across these replays. The second family is evaluated at the generation-consistent 15\,m setting on all seven maps (plus all six conditions on env1). In the primary family, two to four of the eight candidates per map have delivered scores that change with the communication range; the rest are range-invariant.

\textbf{Enumeration.} For every candidate--condition pair we replay all $2^4 = 16$ subsets from fresh worlds, 768 replays per map for the primary family, 128 per map for the second at 15\,m, with the full-team consistency check of Section~\ref{sec:replay} at the 15\,m generation condition. All scores use the same candidate pool, trajectories, and subset replays.\footnote{Analysis code, all subset values, and an interactive results page: \url{https://github.com/williamteo/pairwise-regret}.}

\section{Results}
\label{sec:results}

\subsection{Selection changes across maps and candidate families}
\label{sec:recurrence}

Figure~\ref{fig:env1} shows the two plans selected on env1 at the 15\,m generation range. Both scores rank the same eight logged candidate plans. The candidate selected by $\Ftwo$ has $\Fobj(N) = 0.456$, compared with 0.794 for the plan $\Fobj$ ranks first, giving regret 0.337 of map coverage. The top-two margin under $\Fobj$ is 0.016. The $\Ftwo$-selected team explores 68.3\% of the map and delivers 45.6\%. Of the 33.7 percentage point gap to the exact winner, 22.6 points correspond to explored area that did not reach the base (Fig.~\ref{fig:env1}b). The $\Fobj$-selected plan delivers everything it explores.

\begin{figure}[t]
\centering
\includegraphics[width=\columnwidth]{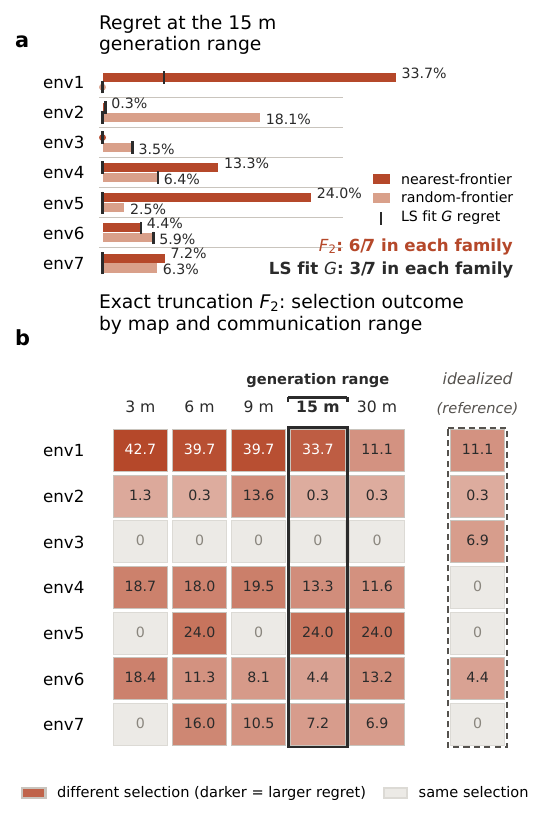}
\caption{\textbf{Pairwise-score regret across the seven benchmark maps.}
(a)~At the 15\,m generation range, exact order-2 truncation has nonzero regret on six of seven maps in each candidate family; the least-squares (LS) two-additive fit $\Gfit$ has nonzero regret on three of seven in each. Bars show $\Ftwo$ regret for the two families; charcoal ticks mark $\Gfit$'s regret, with a tick at zero where the fit selects $\Fobj$'s plan; a dot at zero in the family's bar color marks maps where $\Ftwo$ does the same.
(b)~Exact-truncation ($\Ftwo$) outcomes for the nearest-frontier pool across communication ranges; accented cells mark conditions where $\Fobj$ and $\Ftwo$ select different plans, darker for larger regret (printed as \% of map coverage); neutral cells select the same
plan. The boxed column is the generation range; the idealized column removes the communication constraint as a reference.}
\label{fig:recurrence}
\vspace{-12pt}
\end{figure}

Figure~\ref{fig:recurrence} reports the same comparison on all seven released maps. At the 15\,m generation range, $\Ftwo$ selects a lower-coverage plan on six of the seven maps in the primary family, with regret from 0.0034 (env2) to 0.337 (env1). Changing the communication range can create or remove these ranking changes. On env5, regret is zero at 3 and 9\,m and 0.240 at 6, 15, and 30\,m. The env5 exact winner has the same scores at every range, and the plan $\Ftwo$ selects at 6, 15, and 30\,m is one of the two env5 candidates whose scores change with range. Its $\Ftwo(N)$ moves above and below the winner's as the range changes. The largest regret at any finite range is 0.427 (env1 at 3\,m). On env3, $\Ftwo$ selects the same plan as $\Fobj$ at every finite range, and a lower-coverage plan under idealized communication.

Ranking the second family's eight uniform-random frontier plans by $\Ftwo$ at 15\,m selects a lower-coverage plan on six of seven maps, with regret from 0.025 (env5) to 0.181 (env2) (Fig.~\ref{fig:recurrence}a). The map with zero regret at 15\,m differs between families: env3 in the primary family and env1 in the second.

\begin{table}[t]
\caption{Selection regret at the 15\,m generation range as \% of map coverage, for the additive score $F_1$, the two pairwise scores, and a uniformly random pick from the eight candidates (expected regret).}
\label{tab:baselines}
\centering
\footnotesize
\setlength{\tabcolsep}{3pt}
\begin{tabular}{l rrrr rrrr}
\toprule
& \multicolumn{4}{c}{Nearest-frontier} & \multicolumn{4}{c}{Random-frontier} \\
\cmidrule(lr){2-5} \cmidrule(lr){6-9}
Map & $F_1$ & $F_2$ & $G$ & Rand. & $F_1$ & $F_2$ & $G$ & Rand. \\
\midrule
env1 & 0 & 33.7 & 7.1 & 24.9 & 15.6 & 0 & 0 & 23.7 \\
env2 & 0 & 0.3 & 0.3 & 7.3 & 1.3 & 18.1 & 0 & 12.4 \\
env3 & 36.6 & 0 & 0 & 25.8 & 0 & 3.5 & 3.5 & 10.1 \\
env4 & 0 & 13.3 & 0 & 23.3 & 0 & 6.4 & 6.4 & 14.0 \\
env5 & 0 & 24.0 & 0 & 22.3 & 3.5 & 2.5 & 0 & 8.7 \\
env6 & 0 & 4.4 & 4.4 & 14.6 & 0 & 5.9 & 5.9 & 9.1 \\
env7 & 0 & 7.2 & 0 & 13.3 & 0 & 6.3 & 0 & 7.5 \\
\bottomrule
\end{tabular}

\end{table}

Table~\ref{tab:baselines} adds the additive score $F_1$ and the expected regret of a uniformly random pick from the eight candidates. At 15\,m, $F_1$ selects the exact winner on six of seven maps in the primary family and four of seven in the second. $\Ftwo$ has larger regret than $F_1$ on 11 of the 14 map--family cases and smaller regret on three (env3 in the primary family, env1 and env5 in the second). On env1 and env5 in the primary family and env2 in the second, the $\Ftwo$ regret exceeds the expected regret of a random pick. On env1 in the primary family and env3 in the second, $\Ftwo$ ranks the exact winner last of the eight candidates.

\begin{figure}[t]
\centering
\includegraphics[width=\columnwidth]{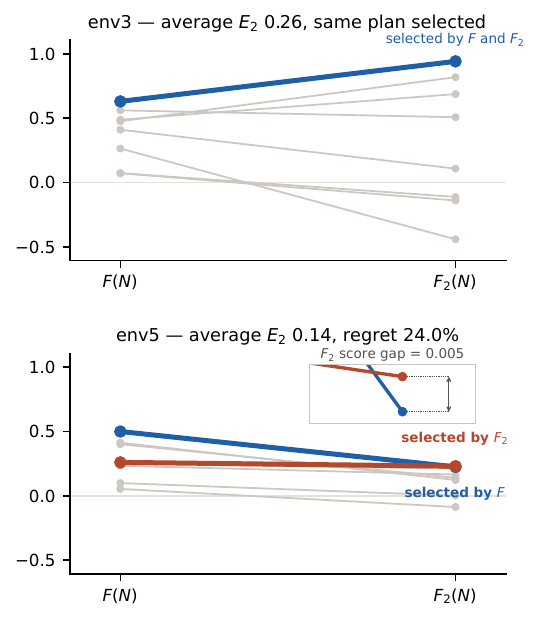}
\caption{\textbf{Exact and order-2 full-team scores for the eight candidates on env3 (top) and env5 (bottom) at 15\,m.} Each line connects a candidate's $\Fobj(N)$ to its $\Ftwo(N)$; truncated scores can be negative (Section~\ref{sec:formulation}). On env3, the average reconstruction error $E_2$ (Eq.~\eqref{eq:e2}) is 0.26 and both scores rank the same plan first. On env5, average $E_2$ is 0.14, the smallest at 15\,m, yet $\Ftwo$ selects a different plan, which delivers 24.0 percentage points less map coverage.}
\label{fig:e2}
\vspace{-12pt}
\end{figure}

\subsection{Reconstruction error and selection regret}
\label{sec:e2null}

Figure~\ref{fig:e2} shows the eight full-team scores on env3 and env5 at 15\,m. The average $E_2$ on env3 (0.26) is nearly double env5's (0.14, the smallest at 15\,m). On env3, $\Fobj$ and $\Ftwo$ rank the same plan first, while on env5 they select different plans and the $\Ftwo$-selected plan delivers 0.240 less coverage; $\Ftwo$ prefers that plan by 0.005. Lower average reconstruction error does not guarantee lower selection regret. The map with the largest average $E_2$ at 15\,m in the primary family, env2 at 0.30, has regret 0.003.

\subsection{Why the ranking changes}
\label{sec:mechanism}

At 15\,m, the top-two margin under $\Fobj$ spans 0.001--0.124 across maps in the primary family and 0.002--0.133 in the second. Full-team truncation errors are larger and differ across candidates: $e_b$ reaches $+1.80$, while $\Fobj$ is bounded by 1. In 10 of the 12 misranked map--family cases at 15\,m the regret exceeds the top-two margin, so the $\Ftwo$-selected plan is not the exact runner-up.

The env1 selection change in Fig.~\ref{fig:env1} gives one numerical example. For the plan that~$\Fobj$ ranks first, the signed order contributions to $\Fobj(N)$ are $+2.50$, $-3.09$, $+1.77$, and $-0.39$, so $\Ftwo(N) = -0.59$ while the complete sum gives $\Fobj(N) = 0.79$, a truncation error of 1.38. For the plan that~$\Ftwo$ ranks first, $\Ftwo(N) = 0.408$ and $\Fobj(N) = 0.456$, an error of 0.049. The two errors differ by 1.33 and the exact-score gap is 0.34, so Eq.~\eqref{eq:condition} holds and the ranking changes.

Under idealized communication, $\Fobj$ is a coverage function. For a coverage function, $\Fobj - \Ftwo$ equals the area seen by exactly three robots plus three times the area seen by all four, normalized by map area. Therefore, greater three- and four-robot overlap increases the amount by which $\Ftwo$ underestimates $\Fobj$. The Bonferroni inequalities give $F_1 \ge \Fobj \ge \Ftwo$ and $F_3 \ge \Fobj$. In the idealized condition every candidate's truncation error $e_b$ is positive. However, under finite communication, $\Fobj$ need not be a coverage function and $e_b$ can be negative, which occurs in 32 of the 280 finite-range primary evaluations. The env1 winner's order sums above show the same alternation, and its four singleton values sum to 2.50 while the team delivers 0.79.

$\Ftwo$ is an algebraic truncation, so its values can fall outside the coverage range $[0,1]$. This occurs in 185 of 336 primary-family evaluations and 79 of 96 second-family evaluations. All fitted full-team scores $\Gfit(N)$ remain within $[0,1]$.

\subsection{Least-squares fitting and order-3 truncation}
\label{sec:fitted}

We also refit the node and edge terms to all subset values. The fitted score $\Gfit$ reduces misranking but does not eliminate it. Across all 54 evaluated settings, $\Gfit$ never has greater regret than $\Ftwo$, and has smaller regret in 28. Across the 42 primary settings, it reduces the number of selection changes from 31 to 10 and removes the largest truncation regrets entirely (env5 at 15\,m, $0.240 \rightarrow 0$; env1 at 3\,m, $0.427 \rightarrow 0$).

At the 15\,m generation range, $\Gfit$ selects a lower-coverage plan on three of seven maps in each family: env1 (regret 0.071, versus 0.337 under exact truncation), env2 (0.003), and env6 (0.044) for the nearest-frontier pool, and env3 (0.035), env4 (0.064), and env6 (0.059) for the random-frontier pool. In the random-frontier pool, these three regrets are the same as the exact truncation's regrets on those maps.

The order-3 truncation $F_3$ includes all triple terms. It removes 30 of $\Ftwo$'s 31 primary selection changes, but introduces four on env3, with regret up to 0.070. In the second family it introduces changes on env1 at 15 and 30\,m, where $\Ftwo$ and $\Gfit$ both agree with $\Fobj$, and at 15\,m it selects a lower-coverage plan on three of seven maps, with regret up to 0.156. Adding order-3 terms removes most $\Ftwo$ selection changes, but it also creates new ones in some settings where $\Ftwo$ agrees with $\Fobj$. At 15\,m, $F_1$ and $F_3$ each change the selection on one map in the primary family and three in the second, against six of seven for $\Ftwo$ in each.

\section{Discussion and Limitations}
\label{sec:discussion}

\textbf{Communication.} The ranking changes occur at finite communication ranges, including the 15\,m generation range. Under idealized communication $\Fobj$ is a coverage function, and its truncation error is the multi-robot overlap of Section~\ref{sec:mechanism}, positive for every candidate on every map. At finite ranges the error can be negative.

\textbf{Limitations.} We evaluate one benchmark, one fixed start configuration, and four-robot teams. Exhaustive subset evaluation grows exponentially with team size, so four robots allow all 16 subsets to be replayed. The plans are deterministic and evaluated by frozen replay of logged trajectories. Removing a robot does not replan the remaining trajectories, so the measured set function is specific to frozen replay. We do not test other team sizes, start poses, or benchmarks.

Both candidate families are eight-plan sets built on the benchmark's frontier machinery and relay-parameter grid, generated at 15\,m. In the primary family, $\Fobj$ ranks the same parameter setting first in every map and condition. The six communication conditions on each map therefore reuse the same map-specific candidate plan. In the second family, the top-ranked plan varies across maps.

We fit one model class: the equal-weight least-squares two-additive model of Eq.~\eqref{eq:gfit}, fitted in-sample. The pairwise terms studied here are set-function terms over robot subsets rather than payoff functions over joint actions, so the results do not transfer directly to coordination-graph methods. The selectors rank a fixed candidate pool; we do not evaluate planners that search or optimize over a pairwise model, and we do not evaluate online replanning.

\section{Conclusion}

For four-robot teams on the seven evaluated maps, both exact order-2 truncation and a fitted two-additive score can select a lower-coverage plan from a fixed pool. Refitting reduces the observed regret but does not eliminate it. Adding pairwise terms to the additive score also does not consistently improve selection.

These results suggest that we should evaluate approximate objectives both by the plans they select and by how accurately they reconstruct objective values. In the tested candidate pools, retaining more interaction terms did not consistently improve selection. Measuring the delivered coverage of the selected plan reveals losses that average reconstruction error alone does not capture.

\IEEEtriggeratref{11}
\bibliographystyle{IEEEtran}
\bibliography{references}

\end{document}